\documentclass[10pt, a4paper]{article}

\usepackage[]{lrec-coling2024} 

\usepackage{natbib}
\usepackage{amsmath}
\usepackage{booktabs}
\usepackage{amssymb}
\usepackage{url}
\usepackage{enumitem}
\usepackage{hyperref}
\usepackage{pifont}

\usepackage{graphicx}
\usepackage{tabularx}
\usepackage{soul}
\usepackage{multirow}
\usepackage{capt-of}

\usepackage{xstring}

\usepackage{color}
\usepackage{cuted}

\newcommand\blfootnote[1]{%
  \begingroup
  \renewcommand\thefootnote{}\footnote{#1}%
  \addtocounter{footnote}{-1}%
  \endgroup
}

\newcommand*{\img}[1]{%
    \raisebox{-.15\baselineskip}{%
        \includegraphics[
        height=\baselineskip,
        width=\baselineskip,
        keepaspectratio,
        ]{#1}%
    }%
}
\newcommand{\olvit}{$\mathbb{OLV}$i$\mathbb{T}$}

\definecolor{Gray}{gray}{0.9}
\definecolor{mygreen}{RGB}{0,110,0}
\definecolor{myblue}{RGB}{0,0,0}
\definecolor{myorange}{RGB}{255,227,200}
\definecolor{mygray}{RGB}{231,231,231}
\newcommand{\cmark}{\ding{51}}
\newcommand{\xmark}{\ding{55}}
\DeclareMathOperator*{\argmax}{arg\,max}

\title{
\olvit: Multi-Modal State Tracking via Attention-Based Embeddings for Video-Grounded Dialog
}

\name{Adnen Abdessaied, Manuel von Hochmeister*\thanks{* Work conducted while at the University of Stuttgart}, Andreas Bulling} 

\address{University of Stuttgart, Bosch, University of Stuttgart \\
         Germany, Germany, Germany \\
         \{adnen.abdessaied, andreas.bulling\}@vis.uni-stuttgart.de\\
         manuel.vonhochmeister@de.bosch.com
}

\abstract{
    We present the $\mathbb{O}$bject $\mathbb{L}$anguage $\mathbb{V}$ideo $\mathbb{T}$ransformer (\olvit) -- a novel model for video dialog operating over a multi-modal attention-based dialog state tracker.
    Existing video dialog models struggle with questions requiring both spatial and temporal localization within videos, long-term temporal reasoning, and accurate object tracking across multiple dialog turns.
    \olvit\, addresses these challenges by maintaining a global dialog state based on the output of an Object State Tracker (OST) and a Language State Tracker (LST):
    while the OST attends to the most important objects within the video, the LST keeps track of the most important linguistic co-references to previous dialog turns.
    In stark contrast to previous works, our approach is generic by nature and is therefore capable of learning continuous multi-modal dialog state representations of the most relevant objects and rounds. As a result, they can be seamlessly integrated into Large Language Models (LLMs) and offer high flexibility in dealing with different datasets and tasks.
    Evaluations on the challenging DVD (response classification) and SIMMC 2.1 (response generation) datasets show that \olvit\, achieves new state-of-the-art performance across both datasets.
 \\ \newline \Keywords{Multi-Modal Learning, Video Dialog, Dialog State Tracking} 
}

\begin{document}

\maketitleabstract

\section{Introduction}

The potential     
\blfootnote{\img{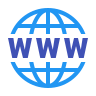} Our project web-page is accessible \href{https://perceptualui.org/publications/abdessaied24_coling/}{here}} of deep learning for tackling unique challenges at the intersection of computer vision (CV) and natural language processing (NLP) has been demonstrated for a wide range of tasks \cite{karpathy2015deep,vqa,visdial,vo2019composing,nokey}.
Among these tasks, video dialog is considered to be one of the most challenging.
In contrast to visual~\cite{vqa} and video~\cite{7780940} question answering, which only require reasoning about a single question, video dialog models have to reason over the whole dialog history.
Furthermore, in contrast to visual dialog~\cite{visdial, Abdessaied_2024_WACV}, video dialog involves reasoning over a dynamic visual input (video) instead of a static image.

While recent video dialog models have improved performance ~\cite{8682583,le-etal-2019-multimodal,pdc}, these gains have largely been marginal, most likely due to the significant challenges posed by this novel task.
Current models suffer from several specific limitations: They struggle with questions that require spatial and temporal localization within the video, they suffer from a general inability of long-term reasoning, and they fail to accurately track objects across multiple dialog turns.
Moreover, they have only been evaluated on benchmarks that were not explicitly designed to minimize biases and test for higher-order reasoning capabilities \cite{8682583,ijcai2018p513}.
To address these limitations, we propose \olvit -- the Object Language Video Transformer for video-grounded dialog.
At the core of \olvit\, are two novel components: an object state tracker (OST) and a language state tracker (LST).
\begin{figure}[!t]
    \begin{minipage}{1\linewidth}
        \centering
        \scalebox{0.97}[0.97]{
            \includegraphics[width=\textwidth]{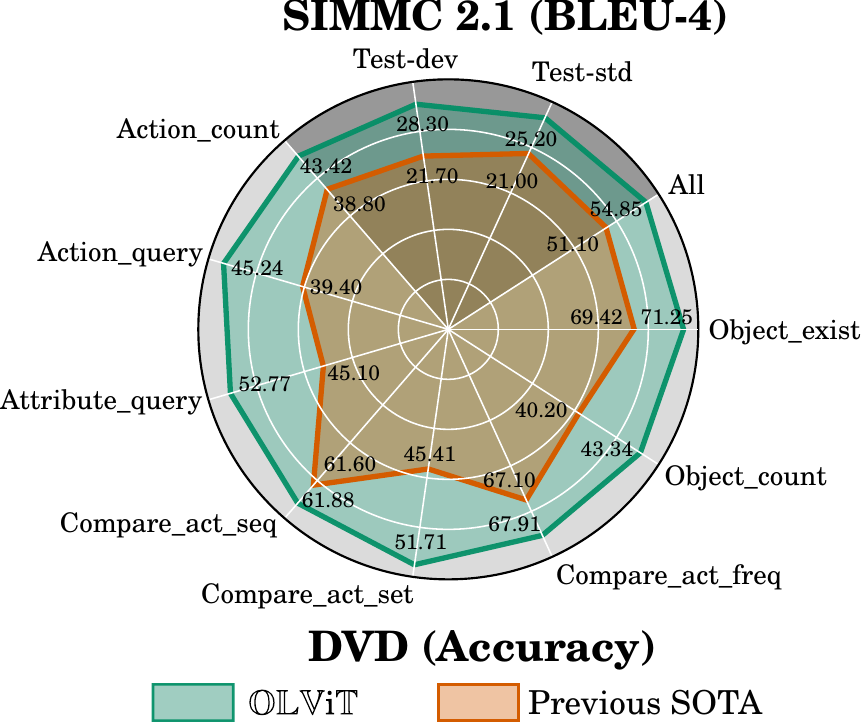}
        }
        \caption{
        \olvit\, outperforms strong baselines and achieves new state-of-the-art results on DVD (classification) and SIMMC 2.1 (generation).
        }
        \label{fig:teaser}
    \end{minipage}
\end{figure}
While the former actively attends to the most relevant objects across different video frames in order to answer the question at hand, the latter encodes the most important linguistic features for a more efficient co-reference resolution (object mentions, attributes, spatial and temporal cues, etc.) across dialog turns.
After each turn, both trackers compute continuous object and language state vectors that are used to update a global dialog state maintained over the course of the entire dialog.
We study different ways of integrating the state vectors in an end-to-end manner within LLMs and demonstrate the applicability of our approach in both a discriminative setting, where the model predicts a full answer, and a generative one where response tokens are predicted one after another. 
In summary, the contributions of our work are three-fold:

\begin{itemize}[wide,labelindent=0cm,leftmargin=0cm]
    \item We introduce the Object Language Video Transformer (\olvit) -- a novel video dialog model that alleviates key limitations of current methods, specifically joint spatial and temporal localization in videos, long-term reasoning, and accurate object tracking across dialog turns.
    \item As a key novelty of our approach, we propose two attention-based video dialog state trackers that
    track the most relevant objects in the visual scene and the most important linguistic co-references to previous dialog turns as well as multiple integration schemes of these states into an end-to-end trainable model.
    \item As can be seen in \autoref{fig:teaser}, empirical results show the applicability of our model not only in a discriminative setting but also in a generative one as well.
    Specifically, \olvit\, achieves a \textit{test} accuracy of $\mathbf{54.85\%}$ on DVD -- outperforming the current state of the art by a $\mathbf{3.75\%}$ margin.
    In addition, \olvit\, outperforms strong baselines on the SIMMC 2.1 generative task by reaching BLEU-4 scores of $\mathbf{28.30}$ and $\mathbf{25.20}$ on the \textit{test-dev} and \textit{test-std} splits, respectively.
\end{itemize}

\section{Related Work}

\paragraph{Video Dialog.}
Video dialog has recently emerged as a particularly challenging task at the intersection of vision and language.
In early work, \citeauthor{ijcai2018p513} proposed to follow a hierarchical attention context learning approach and used a multi-stream network for joint video representation.
Subsequently, \citeauthor{8682583} showed that incorporating audio into an end-to-end video dialog attention-based system could improve performance.
More recently, \citeauthor{jin-etal-2019-video} suggested progressively updating query information based on dialog history and video content
while \citeauthor{le-etal-2019-multimodal} proposed the multi-modal transformer network (MTN) -- an approach that attended to inputs from different embedding spaces and fused the multi-modal information into newly combined representations.
Inspired by the success of graph neural networks, \citeauthor{pdc} proposed a framework for discovering information flows among dialog turns through a semantic graph based on lexical components in each question and answer.
However, most of these models have only been trained and assessed on datasets that were not specifically created to test for higher-order reasoning capabilities \cite{8682583,ijcai2018p513}.
It is therefore unclear to what extent these models exploit the biases present in these datasets.

Our work differs in two distinctive ways: (1) We actively track the most relevant objects and textual facts via two separate multi-modal dialog state trackers that can be seamlessly integrated into pre-trained LLMs. (2) We specifically focus on datasets with minimal bias (i.e. DVD and SIMMC 2.1) that contain sufficient information to confidently assess a model's reasoning capabilities.
In stark contrast, previous models mainly focused on the AVSD dataset \cite{8682583} that does not provide detailed annotations for the different types of reasoning over the spatio-temporal space of video, and thus is not suitable to test for models' higher order reasoning capabilities.

\paragraph{Dialog State Tracking.}
Previous works \cite{dst} have formulated the problem of dialog state tracking as a slot filling task \cite{slot_filling}. 
\citeauthor{mrksic-etal-2017-neural} proposed the Neural Belief Tracker (NBF) to detect slot-value pairs representing the user's goal by iterating over all possible pairs.
\citeauthor{NIPS2018_7348} and \citeauthor{rec_sys} suggested creating a state based on the textual and visual data for visual dialog. They proposed a dialog-based image retrieval model that iteratively interacted with a user in natural language.
\citeauthor{Pang2020} proposed an attention-based tracking of visual features to better generate questions for the ``GuessWhat?!'' dataset \cite{strub2017end}.
Most similar to our work is \cite{Le2022} where the slot filling dialog state paradigm \cite{slot_filling} was simply extended to autoregressively predict the visual attributes of objects in plain text.
However, this work still suffers from further limitations. First, it was specifically designed for the DVD dataset. Therefore, it cannot generalize to other datasets and tasks, thus, significantly limiting its applicability.
Second, its training paradigm heavily relies on extra supervision labels such as bounding box coordinates extracted from a fine-tuned Mask-RCNN \cite{maskrcnn} model.
Finally, its state tracking approach did not lead to significant performance improvements on the down-stream video dialog task.
We instead propose a generic tracking approach that separately tracks the most relevant visual objects and linguistic facts in order to learn continuous state representations that can be seamlessly deployed in conjunction with current pre-trained LLMs.

\paragraph{Multi-modal Reasoning.}
Previous works have investigated whether models are capable of reasoning by introducing fully-controllable and bias-free datasets in visual question answering \cite{johnson2017clevr}, video question answering \cite{clevrer}, visual dialog \cite{clevr_dialog}, and video dialog \cite{dvd}.
Neuro-symbolic models \cite{yi2018neural,Mao2019NeuroSymbolic,shi2019visually,Han2019Visual,clevrer,abdessaied2022neuro} have achieved strong performance on these datasets and have often outperformed fully-connectionist approaches.
However, several recent studies have suggested that transformers \cite{transformer} can deal equally well with the challenges posed by such tasks that have traditionally been tackled using symbolic approaches 
\cite{Lample2020Deep,NEURIPS2020_1457c0d6,deepltl}.
Most similar to our work is the \textit{Aloe} model \cite{aloe} that has been proposed for video question answering and, thus, is not capable of performing multi-turn reasoning within a conversational framework. 
Our model differs from \textit{Aloe} in that it uses a novel multi-modal two-stream state tracker specifically geared towards video dialog.
The state trackers allow our model to jointly attend to the most relevant objects and  previous dialog turns for a more efficient co-reference resolution.

\section{Multi-modal Attention-based Dialog State Tracking}

Our \olvit\, model consists of six main components as shown in \autoref{fig:method}:
An \textit{Object Encoder} uses the unsupervised multi-object network (MONet)~\cite{burgess2019monet} to decompose the video frames into multiple masks that are then used to compute the corresponding scene object embeddings.
Complementing the object encoder, a \textit{Text Encoder} uses a DistilRoBERTa model~\cite{Sanh2019DistilBERTAD} to generate a textual embedding of the current question.
The \textit{Object and Language State Trackers} take the object and language state vectors ${s}_o^{(i-1)}$ and ${s}_l^{(i-1)}$ from the previous dialog turn $(i-1)$ as input and generate updated state vectors ${s}_o^{(i)}$ and ${s}_l^{(i)}$.
A \textit{Combiner} merges both state vectors with the special \texttt{[CLS]} token and
the previous object and text embeddings.
Finally, a transformer \textit{Encoder} block applies a sequence of self-attention and normalization operations on top of the combiner's outputs and uses the \texttt{[CLS]} token to predict the correct answer in the discriminative setting. 
For the generative task, a \textit{Decoder} block is added on top of the encoder and is used to predict the answer tokens auto-regressively.

\subsection{Object Encoder}
\begin{figure*}[!t]
    \begin{minipage}{1\linewidth}
        \centering
        \scalebox{0.98}[0.98]{
            \includegraphics[width=\textwidth]{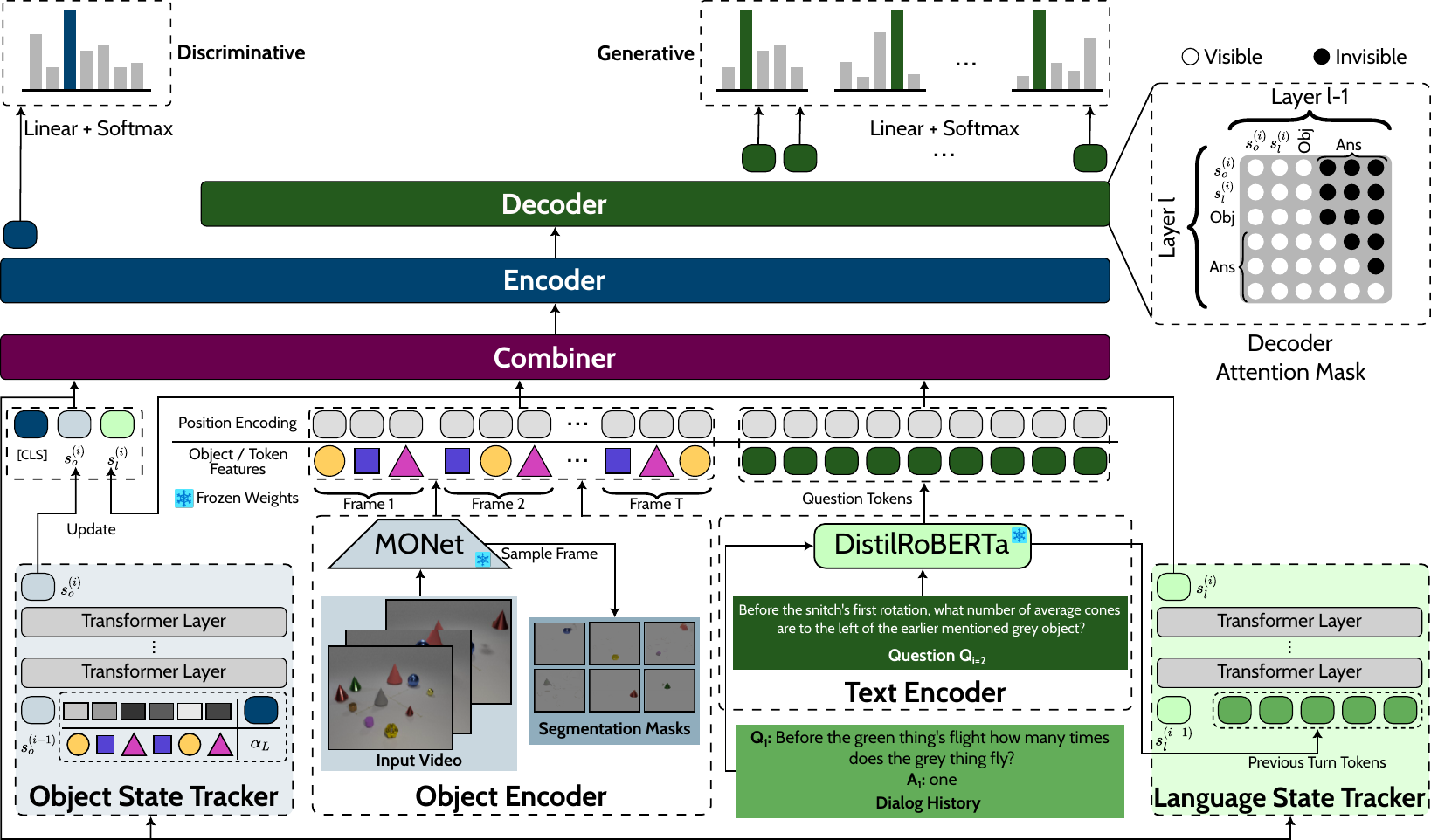}
        }
         \caption{
        Architecture overview of our \olvit\, model.
        It uses MONet and DistilRoBERTa-Base to generate the object embeddings for each frame and the text embeddings, respectively.
        Then, we add position encoding and append the special \texttt{[CLS]} token.
        Finally, we combine the object and language state vectors of the current $i-$th turn ($s_o^{(i)}$ and $s_l^{(i)}$) with the rest of the input, which will be processed by the subsequent transformer layers.
        In the generative setting, a decoder block is added to predict the answer token auto-regressively.
        }
        \label{fig:method}
    \end{minipage}
\end{figure*}

The encoder first samples $T$ frames equidistantly from the video and uses a frozen MONet to obtain $N_o$ object segmentation masks per frame.
Some sample video frames and segmentation masks are illustrated in \autoref{fig:method}.
The masks can be thought of as probabilities of each pixel belonging to a particular object and are encoded in latent variables with mean $\mu_t^n\in\mathbb{R}^{d_{obj}}$, where $n$ indexes the object and $t$ the video frame in which that object occurred.
To obtain object embeddings $H^{obj}_0\in\mathbb{R}^{(TN_o)\times d}$, $\{\mu_t^n\}$ are mapped to the same dimension $d$ as the transformer block using a linear layer.
Position embeddings are added to learn object-frame relationships:

\begin{equation}
    \resizebox{1\hsize}{!}{%
        $H^{obj}_0 = \left[W^{obj}\mu_1^1, .., W^{obj}\mu_1^{N_o}, .., W^{obj}\mu^{N_o}_T\right] + O_{pos},$%
        }
\end{equation}
where $W^{obj}\in \mathbb{R}^{d\times d_{obj}}$ and $O_{pos}\in\mathbb{R}^{(TN_{o})\times d}$ are a learnable parameter and the position embedding.

\subsection{Text Encoder}
The current question at turn $i$ is first tokenized and the resulting $N_w$ sub-word tokens are processed by a frozen pre-trained DistilRoBERTa model~\cite{Sanh2019DistilBERTAD}.
We opted for this approach to improve training efficiency:
We found that a full RoBERTa model \cite{roberta} only marginally improved performance (see Section \ref{sec:full_roberta}).
The $d_w$-dimensional embeddings from the last DistilRoBERTa layer are again mapped to match the input dimension $d$ of the subsequent transformer layers and used as the final word embeddings $H^{w}_0\in\mathbb{R}^{N_w\times d}$, that is: 
\begin{equation}
    H^{w}_0 = \left[W^{w}h_1, ..., W^{w}h_{N_w} \right] + W_{pos},
\end{equation}
where $W^{w}\in \mathbb{R}^{d\times d_{w}}$ is a learnable parameter, $h_j$ is the embedding of the $j$-th token and $W_{pos}\in\mathbb{R}^{N_w\times d}$ is a learnable positional embedding.

\subsection{Dialog State Tracking}
The dialog state tracker is a key novelty of our method.
Its purpose is to track relevant objects within the video over the course of the dialog and to remember co-references to previous dialog turns.
These goals are tackled by two separate sub-components: an object (OST) and a language (LST) state tracker.

\paragraph{Object State Tracker.}
As shown in the bottom left corner of \autoref{fig:method}, the OST takes the object state vector $s_o^{(i-1)}$ and the $k$ most important objects from the previous turn as input.
It consists of $L_{\textrm{ost}}$ transformer layers and uses multi-head self-attention (MSA) with layer normalization (LN) to output an updated $s_o^{(i)}$:
\begin{align}
    H_0^{ost} & = \left[s_o^{(i-1)}, h^{obj}_{L,1}, ..., h^{obj}_{L,k}\right],\\
    h^{obj}_{L, j} & = H_L^{obj}[j, :] \,\,\forall j \in top_k(\alpha_{L}(h_L^{\texttt{[CLS]}}, H^{obj}_L)),\\
    H_l^{ost} & = \textrm{MSA}(\mathrm{LN}(H_{l-1}^{ost})) + H_{l-1}^{ost}, \\
    s_o^{(i)} &= H_{L_{ost}}^{ost}[0, :],
\end{align}
where $L$ is the number of subsequent transformer layers and $\alpha_{L}(h_L^{\texttt{[CLS]}}, H^{obj}_L)\in \mathbb{R}^{TN_o}$ is the attention values between the final embedding of the \texttt{[CLS]} token and the object embeddings.
As such, the updated object state vector $s_o^{(i)}$ holds the most relevant objects for the current turn $i$. 
For the first turn, the object state vector is initialized with a zero vector, i.e.  $s_o^{(0)} = 0$.

For the generative setting where we did not use the \texttt{[CLS]} token in our architecture, we summed the attention values of the object embeddings over all textual tokens in order to determine the most important objects for the given question, i.e. 
\begin{align}
    h^{obj}_{L, j} = H_L^{obj}[j, :] \,\,\forall j \in top_k\left(\alpha_{L}(h_L^{txt}, H^{obj}_L)\right), 
\end{align}
where $h_L^{txt} = \sum_{k=1}^{N_w} h_k^w$ and $\{h_k^w\}$ are the question token embeddings of the last decoder layer.
\begin{figure*}[!t]
    \begin{minipage}{1\linewidth}
        \centering
        \scalebox{0.98}[0.98]{
            \includegraphics[width=\textwidth]{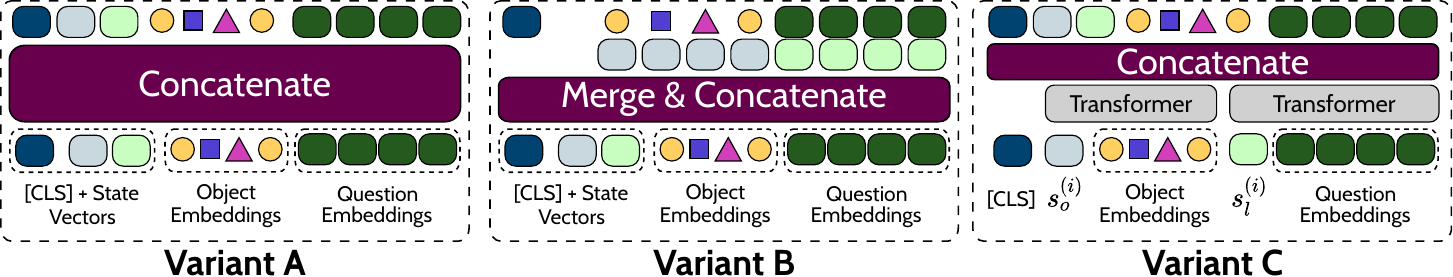}
        }
        \caption{
        Overview of the different variants of our combiner.
        }
        \label{fig:combiner}
    \end{minipage}
\end{figure*}
\paragraph{Language State Tracker.}
As shown in the bottom right corner of \autoref{fig:method}, the LST is a transformer-based module with $L_{lst}$ layers that stores important information previously mentioned in the dialog history which is necessary to resolve potential co-references to previous dialog turns.
The input of its first layer is the previous language state vector $s_l^{(i-1)}$ and turn embeddings.
To calculate the updated language state vector $s_l^{(i)}$ for the current turn $i$, the LST has to be executed on the dialog history comprising of the previous $(i-1)$ turn embeddings.
Each turn is composed of a question-answer pair and its embeddings are generated using the same frozen DistilRoBERTa model as before, that is:
\begin{align}
    H_{0}^{lst} & = \left[s_l^{(i-1)}, W^{w}_{lst} h^{w}_{1}, .., W^{w}_{lst}h^{w}_{n_{i-1}}\right], \\
    H_{l}^{lst} & = \textrm{MSA}(\mathrm{LN}(H_{l-1}^{lst})) + H_{l-1}^{lst} \\
   s_l^{(i)} &= H_{L_{lst}}^{lst}[0, :],
\end{align}
where $W^{w}_{lst}$ is a learnable parameter, $h^w_j$ is the $j$-th token embedding and $n_i$ is the length of $i$-th dialog turn. 
For the first turn, the language state vector is initialized with a zero vector, i.e.  $s_l^{(0)} = 0$.

\subsection{Combiner}
The combiner is responsible for merging the \texttt{[CLS]} token embedding $h_{\texttt{[CLS]}}$ and the two dialog state vectors with the rest of the multi-modal input, i.e. the object and text embeddings.
We propose three variants of how this merging is performed, as illustrated in \autoref{fig:combiner}.

\paragraph{Variant A.}
This variant concatenates all inputs to form the input of the current dialog turn, i.e.
\begin{equation}
    I_A = \left[h^{\texttt{[CLS]}}, s_o^{(i)}, s_l^{(i)}, H_0^{obj}, H_0^{w}\right].
\end{equation}

\paragraph{Variant B.}
This variant first appends the current object state vector $s_o^{(i)}$ to the object embeddings and the current language state vector $s_l^{(i)}$ to the text embeddings.
It then maps them to the $d$-dimensional space using a linear layer and concatenates all inputs like in Variant A, i.e. 
\begin{align}
    I_B &= \left[h^{\texttt{[CLS]}}, \tilde{H}_0^{obj}, \tilde{H}_0^{w}\right],\\
    \tilde{H}_0^{obj} &=  \left[W^b [s_o^{(i)}; h^{obj}_1], ..., W^b [s_o^{(i)}; h^{obj}_{TN_o}]\right], \\
    \tilde{H}_0^{w} &= \left[W^b [s_l^{(i)}; h_1], ..., W^b [s_l^{(i)}; h_{N_w}]\right],
\end{align}
where $W^b\in\mathbb{R}^{d\times (2d)}$ is a learnable parameter.

\paragraph{Variant C.}
This variant uses two small transformers operating on the concatenation of the object state vector with the object embeddings and the language state vector with the text embeddings, respectively.
Then, it concatenates the outputs of their final layers ($H^{obj}$ and $H^{w}$) with the \texttt{[CLS]} token embeddings and the state vectors, i.e. 
\begin{equation}
    I_C = \left[h^{\texttt{[CLS]}}, s_o^{(i)}, s_l^{(i)}, H^{obj}, H^{w}\right].
\end{equation}

\subsection{Encoder/Decoder}
The last component of our model, the encoder, consists of $L$ transformer layers.
It takes the output of the combiner as input and applies multi-head self-attention operations with layer normalization.
The output of its final layer is used to either predict or generate the answer to the current question.

\paragraph{Prediction.}
We use a linear layer with softmax to map the final $d$-dimensional \texttt{[CLS]} token embeddings to the $N$-dimensional answer space.
We train our model end-to-end using cross-entropy loss.
During testing, we choose the answer $\hat{a}$ with the highest score, that is:
\begin{equation}
    \hat{a} = \argmax_{a\in A} \left[ log P\left(a | s_o^{(i)}, s_l^{(i)}, H^{obj}, H^{w} \right) \right],
\end{equation}
where $A$ is the set of all candidate answers.

\paragraph{Generation.}
For answer generation, we couple the encoder with a decoder (with the same number of transformer layers $L$ and attention heads) and append the ground truth answer to the question.
We then train the model end-to-end using the teacher forcing strategy \cite{williams1989learning} while making sure that only the left part is visible to each answer token when attention is applied as shown in the attention mask of \autoref{fig:method}.
While testing, we select the token $\hat{y}_{j}$ at the current step $j$  with the highest score until the \texttt{[EOS]} token is predicted or if a maximum length of $40$ tokens is reached, i.e.
\begin{equation}
    \resizebox{1\hsize}{!}{%
        $\hat{y}_{j} = \displaystyle \argmax_{y\in V} \left[log P\left(y | \hat{Y}_{j-1}, s_o^{(i)},  s_l^{(i)}, H^{obj}, H^{w} \right)\right],$%
        }
\end{equation}
where $\hat{Y}_{j-1} = [\hat{y}_{1}, ..., \hat{y}_{j-1}]$ is the set of previously predicted tokens and $V$ is the vocabulary.

\section{Experiments}
\paragraph{Datasets \& Metrics.}
To evaluate the performance of our model for both discriminative and generative task settings we used the DVD \cite{dvd} and SIMMC 2.1 \cite{simmc} benchmark datasets.
DVD was recently proposed with the goal of assessing higher-order spatio-temporal reasoning capabilities of video dialog models.
It is based on $11$k videos from the challenging CATER dataset \cite{cater} and contains over $100$k dialogs and $1$M question-answer pairs with detailed spario-temporal annotations.
SIMMC 2.1 is a task-oriented dataset that was proposed for realistic virtual assistance scenarios.
It contains $11$k dialogs from the shopping domain.
We used accuracy and BLEU-4 scores to assess the performance of our model on DVD and SIMMC 2.1, respectively.

\begin{figure}[!t]
    \begin{minipage}{1\linewidth}
        \centering
        \scalebox{.8}[.8]{
            \includegraphics[width=\textwidth]{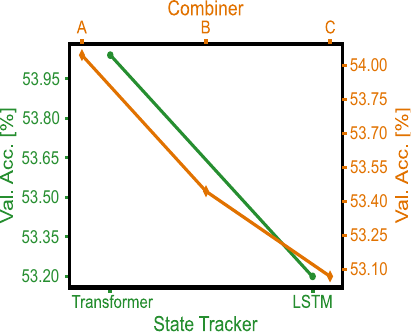}
        }
        \caption{
        Performance comparison of \olvit\, with different combiners and state tracker variants.
        }
        \label{fig:tracker_comb}
    \end{minipage}
\end{figure}

\paragraph{Combiner.}
As can be seen from \autoref{fig:tracker_comb}, variant A of the combiner -- despite having the simplest architecture -- outperformed all other variants with a validation accuracy of $54.01\%$ on DVD.
We hypothesize that this is because concatenating the states vectors to all respective embeddings (variant B) prevents the model from applying attention over the raw state vectors that contain rich information about the previous relevant objects and dialog turns. On the other hand, using additional transformer layers increases complexity and the risk of over-fitting (variant C).
We also experimented with an LSTM-based state tracker which reached a validation accuracy of $53.20\%$ compared to $54.01\%$ achieved by the transformer-based state trackers.

\paragraph{Number of Objects and History Turns.}

\begin{figure}[!t]
    \begin{minipage}{1\linewidth}
        \centering
        \scalebox{.75}[.75]{
            \includegraphics[width=\textwidth]{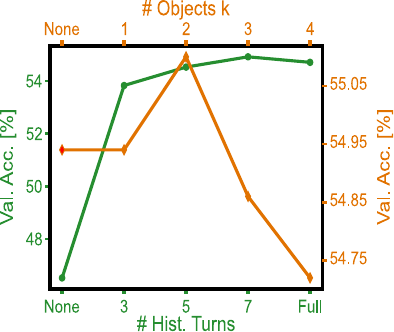}
        }
        \caption{
        Performance comparison of \olvit\, with different numbers of objects and history turns.
        }
        \label{fig:k_h}
    \end{minipage}
\end{figure}

To find the best hyper-parameters (number of the most important previous objects and history turns) of the state tracker variant A, we first used the full history, i.e. all previous turns, and varied the number of objects.
Similarly to the previous experiment, we considered the two variables independently to keep the size of the search space tractable.
Once the optimal number of objects was found in the OST, we optimized the number of previous turns in the LST.
As can be seen from \autoref{fig:k_h}, we increased the number of object embeddings in each OST step (i.e. None, 1, 2, 3, and 4) while keeping the entire dialog history.
\olvit\, achieved its peak validation accuracy of $55.10\%$ when using two object embeddings.
With more embeddings, performance started to decrease and reached $54.86\%$ and $54.72\%$ with three and four object embeddings, respectively.
We then fixed the number of object embeddings to two and varied the history length.
The best validation accuracy of $55.39\%$ was reached when we used a history of seven previous turns.
Using less or more turns resulted in reduced performance, i.e. $54.53\%$ and $54.71\%$ with $10$ and $5$ turns, respectively.
These optimal values were then fixed for the rest of the experiments unless it was explicitly stated otherwise.

\paragraph{Baselines.} We compared \olvit\, using the previous optimal values against strong baselines that (1) were recently published for video dialog state tracking, i.e. VDTN \cite{Le2022}, and (2) hold the current state-of-the-art performance on both DVD and SIMMC 2.1 datasets, i.e. MTN \cite{le-etal-2019-multimodal} and the GPT-2 \cite{radford2019language} based MM-DST model \cite{moon-etal-2020-situated}.
\begin{table*}[!t]
\centering
\scalebox{0.75}[0.75]{

\begin{tabular}{lccccccccccc}
\toprule
\multirow{2}*{\textbf{Accuracy [\%]}} & \textbf{Answer} & \textbf{Q-type} & \textbf{Q-type} & \multirow{2}*{\textbf{Q-retrieval}} & \textbf{RNN} & \textbf{HRNN}& \multicolumn{1}{c}{\textbf{HRNN}} &\textbf{VDTN}$^\dag$ & \textbf{MTN} & \textbf{\olvit} & \multirow{2}*{$\Delta$ \textbf{[\%]}}\\

\cmidrule(r){8-11}

 & \textbf{Prior} & \textbf{(Random)} & \textbf{(Freq)} &  & \textbf{(Q)} &\textbf{(C+Q)}& \multicolumn{4}{c}{\textbf{(C+Q+V)}}&    \\
\midrule
\textbf{Action count}       & $0.0$   & $9.30$  & $23.40$ & $19.80$ & $16.30$ & $28.20$ & $36.00$               &  $38.78$             & ${\underline{38.80}}$ & ${\mathbf{43.42}}$ & \color{mygreen}$\mathbf{+4.62}$ \\
\textbf{Action query}         & $0.0$   & $12.70$ & $23.70$ & $20.60$ & $25.80$ & $33.10$ & $38.60$               &  $39.37$             & ${\underline{39.40}}$ & ${\mathbf{45.24}}$ & \color{mygreen}$\mathbf{+5.84}$ \\
\textbf{Attribute query}      & $0.0$   & $32.90$ & $38.70$ & $39.40$ & $38.10$ & $39.20$ & ${\underline{45.10}}$ &  $42.93$             & $43.10$             & ${\mathbf{52.77}}$ & \color{mygreen}$\mathbf{+7.67}$ \\
\textbf{Compare action seq}   & $33.40$ & $34.10$ & $37.30$ & $35.10$ & $45.50$ & $52.50$ & $57.50$               &  $61.57$             & ${\underline{61.60}}$ & ${\mathbf{61.88}}$ & \color{mygreen}$\mathbf{+0.28}$ \\
\textbf{Compare action set}   & $25.10$ & $28.20$ & $36.30$ & $28.20$ & $32.80$ & $40.00$ & $44.30$               &  \underline{$45.41$} & ${{45.40}}$ & ${\mathbf{51.71}}$ & \color{mygreen}$\mathbf{+6.30}$ \\
\textbf{Compare action freq}  & $48.50$ & $50.00$ & $50.50$ & $44.40$ & $58.40$ & $56.90$ & $65.20$               &  $66.42$             & ${\underline{67.10}}$ & ${\mathbf{67.91}}$ & \color{mygreen}$\mathbf{+0.81}$ \\
\textbf{Object count}         & $0.0$   & $9.10$  & $23.30$ & $18.80$ & $26.20$ & $38.60$ & ${\underline{40.20}}$ &  $39.86$             & $39.90$             & ${\mathbf{43.34}}$ & \color{mygreen}$\mathbf{+3.14}$ \\
\textbf{Object exist}         & $48.90$ & $49.80$ & $51.10$ & $54.40$ & $66.40$ & $67.00$ & ${{69.40}}$           &  \underline{$69.42$} & $69.00$             & ${\mathbf{71.25}}$ & \color{mygreen}$\mathbf{+1.83}$ \\
\midrule
\textbf{None}                 & $0.0$   & $32.10$ & $38.30$ & $39.00$ & $38.30$ & $39.50$ & ${\underline{45.10}}$ & $43.51$             & $43.40$             & ${\mathbf{52.74}}$ & \color{mygreen}$\mathbf{+7.64}$\\
\textbf{Atomic (non-spatial)} & $18.80$ & $26.30$ & $31.90$ & $42.40$ & $47.20$ & $47.80$ & ${\underline{50.70}}$ & $48.88$             & $48.90$             & ${\mathbf{56.54}}$ & \color{mygreen}$\mathbf{+5.84}$\\
\textbf{Atomic (spatial)}     & $21.20$ & $27.30$ & $35.50$ & $27.60$ & $36.80$ & $46.00$ & ${\underline{47.60}}$ & $47.12$             & $47.10$             & ${\mathbf{49.61}}$ & \color{mygreen}$\mathbf{+2.01}$\\
\textbf{Compositional}        & $22.80$ & $28.00$ & $35.40$ & $32.10$ & $40.00$ & $45.80$ & $51.40$               & $53.18$             & ${\underline{53.20}}$ & ${\mathbf{56.70}}$ & \color{mygreen}$\mathbf{+3.50}$\\
\midrule
\textbf{Transfer (attribute)} & $0.0$   & $30.70$ & $45.50$ & $37.10$ & $40.80$ & $45.70$ & $57.30$               & \underline{$57.70$} & ${\underline{57.70}}$ & ${\mathbf{61.28}}$ & \color{mygreen}$\mathbf{+3.58}$\\
\textbf{Transfer (spatial)}   & $49.80$ & $42.40$ & $44.90$ & $26.40$ & $29.60$ & ${\underline{48.10}}$ & $47.40$ & $47.86$ & ${48.00}$ & ${\mathbf{50.50}}$ & \color{mygreen}$\mathbf{+2.40}$\\
\textbf{Transfer (temporal)}  & $28.90$ & $38.40$ & $22.60$ & $3.00$ & $30.20$  & $53.50$ & $64.60$               & $68.72$ & ${\underline{69.00}}$ & ${\mathbf{74.83}}$ & \color{mygreen}$\mathbf{+5.83}$\\
\midrule
\textbf{All}                  & $21.30$ & $27.80$ & $35.30$ & $32.10$ & $39.70$ & $45.80$ & $50.20$               &  $51.02$     & ${\underline{51.10}}$ & ${\mathbf{54.85}}$ & \color{mygreen}$\mathbf{+3.75}$ \\
\bottomrule

\end{tabular}
}
\caption{
Performance comparison on DVD \textit{test} split.
Best and second best performances are in \textbf{bold} and \underline{underlined}, respectively.
Q, C, and V denote question, context, and visual input, respectively, and $\dag$ denotes training with additional supervision.}
\label{tab:dvd}
\end{table*}

\begin{table}[!t]
\centering
\scalebox{.65}[.65]{

\begin{tabular}{lcccccc}
\toprule
& \multicolumn{3}{c}{\textbf{test-dev}} & \multicolumn{3}{c}{\textbf{test-std}} \\
\cmidrule(r){2-4} \cmidrule(r){5-7}

& \textbf{MM-DST} & \textbf{MTN} & \textbf{\olvit} & \textbf{MM-DST} & \textbf{MTN} &  \textbf{\olvit}\\
\midrule
\textbf{BLEU-4} & $19.20$ & $\underline{21.70}$ &  ${\mathbf{28.30}}$ & $19.20$ & $\underline{21.00}$ &  ${\mathbf{25.20}}$ \\
\textbf{Rel.} $\Delta$ \textbf{[\%]} &  \color{mygreen}$\mathbf{+47.4}$ &  \color{mygreen}$\mathbf{+30.4}$ &  $-$ & \color{mygreen}$\mathbf{+31.3}$ &  \color{mygreen}$\mathbf{+20.0}$ &  $\mathbf{-}$   \\
\bottomrule

\end{tabular}
}
\caption{Performance comparison on SIMMC 2.1 \textit{test-dev} and \textit{test-std} splits.
}
\label{tab:simmc}
\end{table}

\section{Results}
\subsection{Results on DVD}
\paragraph{Quantitative Analysis.}
We first evaluated \olvit\, on the discriminative video dialog task. Given a video, a dialog history, and a question, the model needs to predict the correct answer from a pool of $N = 40$ candidate answers.
As can be seen from \autoref{tab:dvd}, our model reached an overall test accuracy of $54.85\%$, thereby outperforming the state of the art model by $3.75\%$ absolute points (last column of \autoref{tab:dvd}).
Not only does our model reached a new overall state-of-the-art test accuracy, it did so by improving the performance across \textit{all} question categories.
This is in stark contrast to previous methods (e.g. MTN \cite{le-etal-2019-multimodal} and more importantly the recent VDTN model \cite{Le2022}) that typically only improved performance for a subset of categories.
Moreover, our method performed particularly well on challenging categories that require accurate object tracking and that existing models tend to struggle with.
Specifically, \olvit\, improved the performance of \textit{action count, action query} and \textit{attribute query} by $4.62\%$, $5.84\%$, and $7.67\%$ over the state of art, respectively.
Furthermore, our model outperformed the state of the art across all \textit{transfer} categories (penultimate section of \autoref{tab:dvd}) matching our hypothesis that it has superior spatio-temporal reasoning capabilities with accuracies of $61.28\%$, $50.50\%$, and $74.83\%$ on the \textit{transfer (attribute)}, \textit{transfer (spatial)}, and \textit{transfer (temporal)} categories, respectively.
This corresponds to respective improvements of $3.58\%$, $2.40\%$, and $5.83\%$ over the state of the art.

\begin{figure}[!t]
    \begin{minipage}{1\linewidth}
        \centering
        \scalebox{0.96}[0.96]{
            \includegraphics[width=\textwidth]{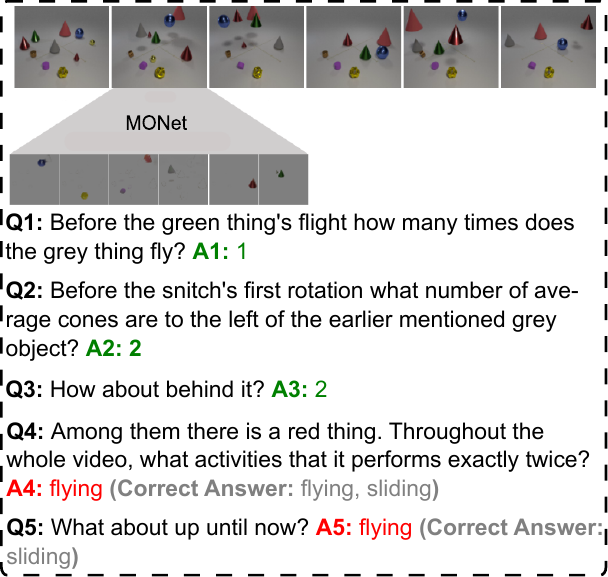}
        }
        \caption{
        \olvit\, predictions on a random DVD sample from the test split.
        }
        \label{fig:qualitative_dvd}
    \end{minipage}
\end{figure}

\paragraph{Qualitative Analysis.}
\autoref{fig:qualitative_dvd} shows sample predictions of \olvit\, on the DVD test split together with some video frames and MONet object masks.
While our model answered the first three questions correctly, it failed at the last two.
For instance, when asked about the actions the red metal cone performs, it predicted ``\textit{flying}'' instead of ``\textit{flying, sliding}''.
This can most likely be attributed to the difficulty of the CATER videos -- deciding between ``\textit{flying}'' and ``\textit{sliding}'' requires the model to reason about the object's shadow which are hard to acquire from the object embeddings.

\subsection{Results on SIMMC 2.1}
For the generative setting, we evaluated \olvit\, on the SIMMC 2.1 dataset that comes with two test splits (\textit{test-dev} and \textit{test-std}).
The ground-truth answers of the former are publicly available but those of the latter are withheld by the creators of the dataset.
As can be seen in \autoref{tab:simmc}, our model outperformed strong baselines on the test-dev split by reaching a BLEU-4 score of $28.30$ compared to $19.20$ and $21.70$ achieved by MM-DST and MTN, respectively.
This corresponds to relative improvements of $47.4\%$ and $30.4\%$, respectively.
\autoref{tab:simmc} also shows the performance of \olvit\, as well as the baselines on the \textit{test-std} split of SIMMC 2.1 where it outperformed both baselines by a considerable margin achieving a BLEU-4 score of $25.20$. This corresponds to relative improvements of $31.3\%$ and $20.0\%$ over MM-DST and MTN, respectively.

\begin{figure}[!t]
    \begin{minipage}{1\linewidth}
        \centering
        \scalebox{0.97}[0.97]{
            \includegraphics[width=\textwidth]{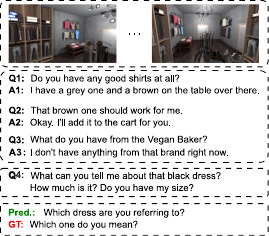}
        }
        \caption{
        \olvit\,  predictions on a random SIMMC 2.1 sample from the test split.
        }
        \label{fig:qualitative_simmc}
    \end{minipage}
\end{figure}

To qualitatively assess the performance of our model, we 
show in \autoref{fig:qualitative_simmc} the generated answer of our model on a randomly sampled example from the SIMMC 2.1 \textit{test-dev} split.
Although there is not a big overlap in the used words between our model's prediction (green) and the ground-truth (red), the two responses are semantically similar and lead to the same action of the virtual assistant, i.e. ask for more information from the user.

\begin{table}[!t]
\centering
\scalebox{0.82}[0.82]{

\begin{tabular}{cccccc}
\toprule
\multicolumn{4}{c}{\textbf{Ablated \olvit}} & \multicolumn{2}{c}{\textbf{State}} \\ 
\cmidrule(r){1-4} 
\multicolumn{2}{c}{\textbf{DVD}} & \multicolumn{2}{c}{\textbf{SIMMC 2.1}} & \multicolumn{2}{c}{\textbf{Trackers}}\\
\cmidrule(r){1-2} \cmidrule(r){3-4} \cmidrule(r){5-6}

 \textbf{Acc. [\%]} & $\Delta$\textbf{[\%]} & \textbf{BLEU-4} & \textbf{Rel.} $\Delta$ \textbf{[\%]} & \textbf{OST} & \textbf{LST} \\
\midrule
$44.11$ & {$11.28$} & $21.50$ & {$22.79$} & \color{red}{\xmark} & \color{red}{\xmark} \\
$46.52^{\ddag}$ & {$8.87$}  &  $21.70^\dag$ & {$21.66$} & \color{blue}{\cmark} & \color{red}{\xmark} \\
$\underline{54.94}^\ddag$ & $0.45$ &  $\underline{26.30}^\ddag$ & {$0.38$} & \color{red}{\xmark} & \color{blue}{\cmark} \\
${\mathbf{55.39}}^\dag$  & $\mathbf{-}$ & ${\mathbf{26.40}^\dag}$ & $\mathbf{-}$ & \color{blue}{\cmark} & \color{blue}{\cmark} \\
\bottomrule
\end{tabular}
}
\caption{Performance comparison of different ablated versions on DVD and SIMMC 2.1 \textit{val} splits. 
$\dag$ and $\ddag$ represent significant improvement with $p < 0.05$ and $p < 0.01$ compared to the second best score in each column, respectively.
}
\label{tab:ablations}
\end{table}

\subsection{Ablation Study}
\label{sec:full_roberta}

\paragraph{Performance without OST \& LST.}
We first evaluated a version of our model without any state trackers (first row in \autoref{tab:ablations}).
It is important to note that in this case, the model still has access to the MONet object embeddings but not to the previous history turns.
This ablated version performed poorly,
only reaching $44.11\%$ validation accuracy on DVD and $21.50$ BLEU-4 score on SIMMC.

\paragraph{Performance with the OST.}
Adding the OST (second row in \autoref{tab:ablations})
resulted in a notable performance improvement on both datasets, i.e. $46.52\%$ on DVD and $21.70$ on SIMMC.
We emphasize that this version still does not have access to the previous history turns and treats the task as a simple single-turn question answering task. 
This finding indicates the positive influence of the OST in helping the model to focus on the most important objects when answering a question.

\begin{table}[!t]
\centering
\scalebox{0.85}[0.85]{
\begin{tabular}{lccc}
\toprule

\multirow{2}*{\textbf{LM}}& \textbf{\# of LM} & \textbf{\# of Layers} & \textbf{Accuracy} \\
                           & \textbf{param.} & $\mathbf{L}$ & \textbf{[\%]} \\
\midrule
None & - & $4$ & $54.46$ \\
None & - & $16$ & $54.54$ \\
\midrule
DistillRoBERTa & $81.5$M & $4$ & $55.39$ \\
RoBERTa & $354$M & $\mathbf{4}$ & $\mathbf{55.41}$ \\
\bottomrule

\end{tabular}
}
\caption{The effect of pre-trained language models on the performance of our model. Performance is measured on the DVD \textit{val} split.}
\label{tab:lm}
\end{table}

\paragraph{Performance with the LST.}
Adding the LST (third row in \autoref{tab:ablations}) lifted the performance on both DVD and SIMMC 2.1 by a considerable margin compared to the version with only the OST.
We note here that this version, in stark contrast to the previous one, not only had access to the visual input in the form of MONet embeddings but also to the previous dialog turns.
This variant reached a validation accuracy of $54.94\%$ and a BLEU-4 score of $26.30$.
However, it still under-performed our full model with both dialog trackers in action which achieved the best validation performance on both datasets, i.e. $55.39\%$ accuracy on DVD and $26.40$ BLEU-4 score on SIMMC 2.1. 

\paragraph{Performance with Pre-trained LLM.}
To assess the effect of pre-trained language models on \olvit, we completely removed the DistillRoBERTa model and trained the encoder layers from scratch on the DVD dataset.
Thereby, we increased the number of the encoder transformer layers from $4$ to $16$ to compensate for the removal of the pre-trained language model.
As can be seen from \autoref{tab:lm}, our model's DVD validation accuracy dropped from $55.41\%$ to $54.46\%$ when DistillRoBERTa was replaced by $4$ encoder layers.
However, its performance improved when we increased the number of these layers to $16$ and reached a validation accuracy of $54.54\%$.
Finally, using a full RoBERTa model only improved the model's accuracy by $0.02\%$.
As a result, we decided to use a DistillRoBERTa model in order to train efficiently.

\section{Conclusion}

In this work, we proposed
\olvit\, -- a novel model for video dialog operating over a novel two-stream multi-modal attention-based dialog state tracker that jointly learns visual object representations and textual embeddings.
Through extensive experiments on two challenging datasets, we demonstrated significant improvements over strong baselines for discriminative and generative tasks. 
Our results are highly encouraging and underline the importance of performing multi-modal dialog state tracking for a more reliable higher-order reasoning of (video) dialog systems.
We strongly believe that real-world multi-modal dialog tasks can equally benefit from our novel multi-modal state tracking approach.
We leave this for future work.

\section*{Acknowledgment}
A. Bulling was funded by the European Research Council (ERC; grant agreement 801708).

\nocite{*}
\section{Bibliographical References}\label{sec:reference}

\bibliographystyle{lrec-coling2024-natbib}
\bibliography{lrec-coling2024-example}

\begin{thebibliography}{49}
\expandafter\ifx\csname natexlab\endcsname\relax\def\natexlab#1{#1}\fi

\bibitem[{Abdessaied et~al.(2022)Abdessaied, B{\^a}ce, and
  Bulling}]{abdessaied2022neuro}
Adnen Abdessaied, Mihai B{\^a}ce, and Andreas Bulling. 2022.
\newblock {Neuro-Symbolic Visual Dialog}.
\newblock In \emph{COLING}.

\bibitem[{Abdessaied et~al.(2024)Abdessaied, Shi, and
  Bulling}]{Abdessaied_2024_WACV}
Adnen Abdessaied, Lei Shi, and Andreas Bulling. 2024.
\newblock {VD-GR: Boosting Visual Dialog With Cascaded Spatial-Temporal
  Multi-Modal Graphs}.
\newblock In \emph{WACV}.

\bibitem[{Antol et~al.(2015)Antol, Agrawal, Lu, Mitchell, Batra, Zitnick, and
  Parikh}]{vqa}
Stanislaw Antol, Aishwarya Agrawal, Jiasen Lu, Margaret Mitchell, Dhruv Batra,
  C.~Lawrence Zitnick, and Devi Parikh. 2015.
\newblock {VQA}: {V}isual {Q}uestion {A}nswering.
\newblock In \emph{ICCV}.

\bibitem[{Brown et~al.(2020)Brown, Mann, Ryder, Subbiah, Kaplan, Dhariwal,
  Neelakantan, Shyam, Sastry, Askell, Agarwal, Herbert-Voss, Krueger, Henighan,
  Child, Ramesh, Ziegler, Wu, Winter, Hesse, Chen, Sigler, Litwin, Gray, Chess,
  Clark, Berner, McCandlish et~al.}]{NEURIPS2020_1457c0d6}
Tom Brown, Benjamin Mann, Nick Ryder, Melanie Subbiah, Jared~D Kaplan, Prafulla
  Dhariwal, Arvind Neelakantan, Pranav Shyam, Girish Sastry, Amanda Askell,
  Sandhini Agarwal, Ariel Herbert-Voss, Gretchen Krueger, Tom Henighan, Rewon
  Child, Aditya Ramesh, Daniel Ziegler, Jeffrey Wu, Clemens Winter, Chris
  Hesse, Mark Chen, Eric Sigler, Mateusz Litwin, Scott Gray, Benjamin Chess,
  Jack Clark, Christopher Berner, Sam McCandlish, et~al. 2020.
\newblock Language models are few-shot learners.
\newblock In \emph{NeurIPS}.

\bibitem[{Burgess et~al.(2019)Burgess, Matthey, Watters, Kabra, Higgins,
  Botvinick, and Lerchner}]{burgess2019monet}
Christopher~P. Burgess, Loic Matthey, Nicholas Watters, Rishabh Kabra, Irina
  Higgins, Matt Botvinick, and Alexander Lerchner. 2019.
\newblock {MONet: Unsupervised Scene Decomposition and Representation}.
\newblock In \emph{arXiv, 1901.11390}.

\bibitem[{Das et~al.(2017)Das, Kottur, Gupta, Singh, Yadav, Moura, Parikh, and
  Batra}]{visdial}
Abhishek Das, Satwik Kottur, Khushi Gupta, Avi Singh, Deshraj Yadav,
  Jos\'e~M.F. Moura, Devi Parikh, and Dhruv Batra. 2017.
\newblock {V}isual {D}ialog.
\newblock In \emph{CVPR}.

\bibitem[{Ding et~al.(2021)Ding, Hill, Santoro, Reynolds, and Botvinick}]{aloe}
David Ding, Felix Hill, Adam Santoro, Malcolm Reynolds, and Matthew Botvinick.
  2021.
\newblock {Attention over Learned Object Embeddings Enables Complex Visual
  Reasoning}.
\newblock In \emph{NeurIPS}.

\bibitem[{Girdhar and Ramanan(2020)}]{cater}
Rohit Girdhar and Deva Ramanan. 2020.
\newblock {CATER: A diagnostic dataset for Compositional Actions and TEmporal
  Reasoning}.
\newblock In \emph{ICLR}.

\bibitem[{Guo et~al.(2018)Guo, Wu, Cheng, Rennie, Tesauro, and
  Feris}]{NIPS2018_7348}
Xiaoxiao Guo, Hui Wu, Yu~Cheng, Steven Rennie, Gerald Tesauro, and Rogerio
  Feris. 2018.
\newblock Dialog-based interactive image retrieval.
\newblock In \emph{NeurIPS}.

\bibitem[{Hahn et~al.(2021)Hahn, Schmitt, Kreber, Rabe, and
  Finkbeiner}]{deepltl}
Christopher Hahn, Frederik Schmitt, Jens~U. Kreber, Markus~Norman Rabe, and
  Bernd Finkbeiner. 2021.
\newblock {Teaching Temporal Logics to Neural Networks}.
\newblock In \emph{ICLR}.

\bibitem[{Han et~al.(2019)Han, Mao, Gan, Tenenbaum, and Wu}]{Han2019Visual}
Chi Han, Jiayuan Mao, Chuang Gan, Joshua~B. Tenenbaum, and Jiajun Wu. 2019.
\newblock {Visual Concept Metaconcept Learning}.
\newblock In \emph{NeurIPS}.

\bibitem[{He et~al.(2017)He, Gkioxari, Dollár, and Girshick}]{maskrcnn}
Kaiming He, Georgia Gkioxari, Piotr Dollár, and Ross Girshick. 2017.
\newblock {Mask R-CNN}.
\newblock In \emph{ICCV}.

\bibitem[{Hochreiter and Schmidhuber(1997)}]{lstm}
Sepp Hochreiter and J\"{u}rgen Schmidhuber. 1997.
\newblock Long short-term memory.
\newblock \emph{Neural Comput.}

\bibitem[{Hori et~al.(2019)Hori, Alamri, Wang, Wichern, Hori, Cherian, Marks,
  Cartillier, Lopes, Das, Essa, Batra, and Parikh}]{8682583}
Chiori Hori, Huda Alamri, Jue Wang, Gordon Wichern, Takaaki Hori, Anoop
  Cherian, Tim~K. Marks, Vincent Cartillier, Raphael~Gontijo Lopes, Abhishek
  Das, Irfan Essa, Dhruv Batra, and Devi Parikh. 2019.
\newblock End-to-end audio visual scene-aware dialog using multimodal
  attention-based video features.
\newblock In \emph{ICASSP}.

\bibitem[{Jin et~al.(2019)Jin, Zhao, Gu, Xiao, Wei, and
  Zhuang}]{jin-etal-2019-video}
Weike Jin, Zhou Zhao, Mao Gu, Jun Xiao, Furu Wei, and Yueting Zhuang. 2019.
\newblock Video dialog via progressive inference and cross-transformer.
\newblock In \emph{EMNLP-IJCNLP}.

\bibitem[{Johnson et~al.(2017)Johnson, Hariharan, van~der Maaten, Fei-Fei,
  Zitnick, and Girshick}]{johnson2017clevr}
Justin Johnson, Bharath Hariharan, Laurens van~der Maaten, Li~Fei-Fei,
  C~Lawrence Zitnick, and Ross Girshick. 2017.
\newblock {CLEVR: A Diagnostic Dataset for Compositional Language and
  Elementary Visual Reasoning}.
\newblock In \emph{CVPR}.

\bibitem[{Karpathy and Fei-Fei(2015)}]{karpathy2015deep}
Andrej Karpathy and Li~Fei-Fei. 2015.
\newblock Deep visual-semantic alignments for generating image descriptions.
\newblock In \emph{CVPR}.

\bibitem[{Kottur et~al.(2021)Kottur, Moon, Geramifard, and Damavandi}]{simmc}
Satwik Kottur, Seungwhan Moon, Alborz Geramifard, and Babak Damavandi. 2021.
\newblock {SIMMC} 2.0: A task-oriented dialog dataset for immersive multimodal
  conversations.
\newblock In \emph{EMNLP}.

\bibitem[{Kottur et~al.(2019)Kottur, Moura, Parikh, Batra, and
  Rohrbach}]{clevr_dialog}
Satwik Kottur, José M.~F. Moura, Devi Parikh, Dhruv Batra, and Marcus
  Rohrbach. 2019.
\newblock {CLEVR-Dialog: A Diagnostic Dataset for Multi-Round Reasoning in
  Visual Dialog}.
\newblock In \emph{NAACL}.

\bibitem[{Lample and Charton(2020)}]{Lample2020Deep}
Guillaume Lample and François Charton. 2020.
\newblock Deep learning for symbolic mathematics.
\newblock In \emph{ICLR}.

\bibitem[{Le et~al.(2021{\natexlab{a}})Le, Chen, and Hoi}]{pdc}
Hung Le, Nancy~F. Chen, and Steven C.~H. Hoi. 2021{\natexlab{a}}.
\newblock Learning reasoning paths over semantic graphs for video-grounded
  dialogues.
\newblock In \emph{ICLR}.

\bibitem[{Le et~al.(2022)Le, Chen, and Hoi}]{Le2022}
Hung Le, Nancy~F. Chen, and Steven~C.H. Hoi. 2022.
\newblock {Multimodal Dialogue State Tracking}.
\newblock In \emph{NAACL}.

\bibitem[{Le et~al.(2019)Le, Sahoo, Chen, and Hoi}]{le-etal-2019-multimodal}
Hung Le, Doyen Sahoo, Nancy Chen, and Steven Hoi. 2019.
\newblock Multimodal transformer networks for end-to-end video-grounded
  dialogue systems.
\newblock In \emph{ACL}.

\bibitem[{Le et~al.(2021{\natexlab{b}})Le, Sankar, Moon, Beirami, Geramifard,
  and Kottur}]{dvd}
Hung Le, Chinnadhurai Sankar, Seungwhan Moon, Ahmad Beirami, Alborz Geramifard,
  and Satwik Kottur. 2021{\natexlab{b}}.
\newblock {DVD}: A diagnostic dataset for multi-step reasoning in video
  grounded dialogue.
\newblock In \emph{ACL}.

\bibitem[{Liu et~al.(2019)Liu, Ott, Goyal, Du, Joshi, Chen, Levy, Lewis,
  Zettlemoyer, and Stoyanov}]{roberta}
Yinhan Liu, Myle Ott, Naman Goyal, Jingfei Du, Mandar Joshi, Danqi Chen, Omer
  Levy, Mike Lewis, Luke Zettlemoyer, and Veselin Stoyanov. 2019.
\newblock {RoBERTa: A Robustly Optimized BERT Pretraining Approach}.
\newblock In \emph{arXiv, 1907.11692}.

\bibitem[{Loshchilov and Hutter(2019)}]{adamw}
Ilya Loshchilov and Frank Hutter. 2019.
\newblock Decoupled weight decay regularization.
\newblock In \emph{ICLR}.

\bibitem[{Mao et~al.(2019)Mao, Gan, Kohli, Tenenbaum, and
  Wu}]{Mao2019NeuroSymbolic}
Jiayuan Mao, Chuang Gan, Pushmeet Kohli, Joshua~B. Tenenbaum, and Jiajun Wu.
  2019.
\newblock {The Neuro-Symbolic Concept Learner: Interpreting Scenes, Words, and
  Sentences From Natural Supervision}.
\newblock In \emph{ICLR}.

\bibitem[{Moon et~al.(2020)Moon, Kottur, Crook, De, Poddar, Levin, Whitney,
  Difranco, Beirami, Cho, Subba, and Geramifard}]{moon-etal-2020-situated}
Seungwhan Moon, Satwik Kottur, Paul Crook, Ankita De, Shivani Poddar, Theodore
  Levin, David Whitney, Daniel Difranco, Ahmad Beirami, Eunjoon Cho, Rajen
  Subba, and Alborz Geramifard. 2020.
\newblock Situated and interactive multimodal conversations.
\newblock In \emph{COLING}.

\bibitem[{Mrk{\v{s}}i{\'c} et~al.(2017)Mrk{\v{s}}i{\'c}, {\'O}~S{\'e}aghdha,
  Wen, Thomson, and Young}]{mrksic-etal-2017-neural}
Nikola Mrk{\v{s}}i{\'c}, Diarmuid {\'O}~S{\'e}aghdha, Tsung-Hsien Wen, Blaise
  Thomson, and Steve Young. 2017.
\newblock Neural belief tracker: Data-driven dialogue state tracking.
\newblock In \emph{ACL}.

\bibitem[{Noroozi et~al.(2020)Noroozi, Zhang, Bakhturina, and Kornuta}]{dst}
Vahid Noroozi, Yang Zhang, Evelina Bakhturina, and Tomasz Kornuta. 2020.
\newblock A fast and robust bert-based dialogue state tracker for schema-guided
  dialogue dataset.
\newblock In \emph{Proceedings of the Workshop on Conversational Systems
  Towards Mainstream Adoption, KDD}.

\bibitem[{Pan et~al.(2020)Pan, Zhang, Ji, and Yang}]{9152761}
Xudong Pan, Mi~Zhang, Shouling Ji, and Min Yang. 2020.
\newblock Privacy risks of general-purpose language models.
\newblock In \emph{Symposium on Security and Privacy}.

\bibitem[{Pang and Wang(2020)}]{Pang2020}
Wei Pang and Xiaojie Wang. 2020.
\newblock {Visual dialogue state tracking for question generation}.
\newblock In \emph{AAAI}.

\bibitem[{Paszke et~al.(2019)Paszke, Gross, Massa, Lerer, Bradbury, Chanan,
  Killeen, Lin, Gimelshein, Antiga, Desmaison, K{\"{o}}pf, Yang, DeVito, Raison
  et~al.}]{pytorch}
Adam Paszke, Sam Gross, Francisco Massa, Adam Lerer, James Bradbury, Gregory
  Chanan, Trevor Killeen, Zeming Lin, Natalia Gimelshein, Luca Antiga, Alban
  Desmaison, Andreas K{\"{o}}pf, Edward~Z. Yang, Zach DeVito, Martin Raison,
  et~al. 2019.
\newblock Pytorch: An imperative style, high-performance deep learning library.
\newblock In \emph{NeurIPS}.

\bibitem[{Radford et~al.(2019)Radford, Wu, Child, Luan, Amodei, and
  Sutskever}]{radford2019language}
Alec Radford, Jeff Wu, Rewon Child, David Luan, Dario Amodei, and Ilya
  Sutskever. 2019.
\newblock Language models are unsupervised multitask learners.
\newblock \emph{OpenAI blog}.

\bibitem[{Rombach et~al.(2022)Rombach, Blattmann, Lorenz, Esser, and
  Ommer}]{nokey}
Robin Rombach, Andreas Blattmann, Dominik Lorenz, Patrick Esser, and Björn
  Ommer. 2022.
\newblock High-resolution image synthesis with latent diffusion models.
\newblock In \emph{CVPR}.

\bibitem[{Sanh et~al.(2019)Sanh, Debut, Chaumond, and
  Wolf}]{Sanh2019DistilBERTAD}
Victor Sanh, Lysandre Debut, Julien Chaumond, and Thomas Wolf. 2019.
\newblock Distilbert, a distilled version of bert: smaller, faster, cheaper and
  lighter.
\newblock In \emph{Proceedings od the Workshop on Energy Efficient Machine
  Learning and Cognitive Computing, NeurIPS}.

\bibitem[{Shi et~al.(2019)Shi, Mao, Gimpel, and Livescu}]{shi2019visually}
Haoyue Shi, Jiayuan Mao, Kevin Gimpel, and Karen Livescu. 2019.
\newblock Visually grounded neural syntax acquisition.
\newblock In \emph{ACL}.

\bibitem[{Song and Raghunathan(2020)}]{risk}
Congzheng Song and Ananth Raghunathan. 2020.
\newblock Information leakage in embedding models.
\newblock In \emph{ACM CCS}.

\bibitem[{Strub et~al.(2017)Strub, De~Vries, Mary, Piot, Courville, and
  Pietquin}]{strub2017end}
Florian Strub, Harm De~Vries, Jeremie Mary, Bilal Piot, Aaron Courville, and
  Olivier Pietquin. 2017.
\newblock End-to-end optimization of goal-driven and visually grounded dialogue
  systems.
\newblock In \emph{IJCAI}.

\bibitem[{Vaswani et~al.(2017)Vaswani, Shazeer, Parmar, Uszkoreit, Jones,
  Gomez, Kaiser, and Polosukhin}]{transformer}
Ashish Vaswani, Noam Shazeer, Niki Parmar, Jakob Uszkoreit, Llion Jones,
  Aidan~N Gomez, \L~ukasz Kaiser, and Illia Polosukhin. 2017.
\newblock {Attention is All you Need}.
\newblock In \emph{NeurIPS}.

\bibitem[{Vo et~al.(2019)Vo, Jiang, Sun, Murphy, Li, Fei-Fei, and
  Hays}]{vo2019composing}
Nam Vo, Lu~Jiang, Chen Sun, Kevin Murphy, Li-Jia Li, Li~Fei-Fei, and James
  Hays. 2019.
\newblock Composing text and image for image retrieval-an empirical odyssey.
\newblock In \emph{CVPR}.

\bibitem[{Williams and Zipser(1989)}]{williams1989learning}
Ronald~J Williams and David Zipser. 1989.
\newblock A learning algorithm for continually running fully recurrent neural
  networks.
\newblock \emph{Neural Comput.}

\bibitem[{Wu et~al.(2022)Wu, Macdonald, and Ounis}]{rec_sys}
Yaxiong Wu, Craig Macdonald, and Iadh Ounis. 2022.
\newblock {Multi-Modal Dialog State Tracking for Interactive Fashion
  Recommendation}.
\newblock In \emph{ACM RecSys}.

\bibitem[{Xu et~al.(2016)Xu, Mei, Yao, and Rui}]{7780940}
Jun Xu, Tao Mei, Ting Yao, and Yong Rui. 2016.
\newblock Msr-vtt: A large video description dataset for bridging video and
  language.
\newblock In \emph{CVPR}.

\bibitem[{Xu and Hu(2018)}]{slot_filling}
Puyang Xu and Qi~Hu. 2018.
\newblock {An End-to-end Approach for Handling Unknown Slot Values in Dialogue
  State Tracking}.
\newblock In \emph{ACL-findings}.

\bibitem[{Yi et~al.(2020)Yi, Gan, Li, Kohli, Wu, Torralba, and
  Tenenbaum}]{clevrer}
Kexin Yi, Chuang Gan, Yunzhu Li, Pushmeet Kohli, Jiajun Wu, Antonio Torralba,
  and Joshua~B. Tenenbaum. 2020.
\newblock {CLEVRER: Collision Events for Video Representation and Reasoning}.
\newblock In \emph{ICLR}.

\bibitem[{Yi et~al.(2018)Yi, Wu, Gan, Torralba, Kohli, and
  Tenenbaum}]{yi2018neural}
Kexin Yi, Jiajun Wu, Chuang Gan, Antonio Torralba, Pushmeet Kohli, and
  Joshua~B. Tenenbaum. 2018.
\newblock Neural-symbolic vqa: Disentangling reasoning from vision and language
  understanding.
\newblock In \emph{NeurIPS}.

\bibitem[{Zhao et~al.(2018)Zhao, Jiang, Cai, Xiao, He, and Pu}]{ijcai2018p513}
Zhou Zhao, Xinghua Jiang, Deng Cai, Jun Xiao, Xiaofei He, and Shiliang Pu.
  2018.
\newblock Multi-turn video question answering via multi-stream hierarchical
  attention context network.
\newblock In \emph{IJCAI}.

\bibitem[{Zhu et~al.(2019)Zhu, Liu, and Han}]{zhu2019deep}
Ligeng Zhu, Zhijian Liu, and Song Han. 2019.
\newblock Deep leakage from gradients.
\newblock In \emph{NeurIPS}.

\end{thebibliography}

\end{document}